\documentclass[twocolumn]{svjour3}
\usepackage{cite}
\usepackage[pdftex]{graphicx}
\usepackage[caption=false,font=footnotesize,labelfont=sf,textfont=sf]{subfig}
\usepackage[cmex10]{amsmath}
\usepackage{url}
\usepackage{rotating}
\usepackage{xcolor}
\usepackage[T1]{fontenc}
\usepackage[utf8]{inputenc}
\usepackage{xspace}
\usepackage{fp}
\usepackage[pdftex]{hyperref}
\usepackage{footnote}
\makesavenoteenv{tabular}
\makesavenoteenv{table}

\newcommand{\num}[1]{#1\xspace}

\newcommand{\IBMUB}{IBM-UB-1\xspace}
\newcommand{\iamondb}{IAM-OnDB\xspace}
\newcommand{\Bezier}{B\'ezier\xspace}

\begin{document}
\title{Fast Multi-language LSTM-based Online Handwriting Recognition}

\author{Victor Carbune \and 
        Pedro Gonnet \and
        Thomas Deselaers \and \\
        Henry A. Rowley \and
        Alexander Daryin \and
        Marcos Calvo \and
        Li-Lun Wang \and
        Daniel Keysers \and
        Sandro Feuz \and
        Philippe Gervais}

\institute{All authors are with Google AI Perception\\
\email{\{vcarbune, gonnet, deselaers, har, shurick, marcoscalvo, llwang, keysers, sfeuz, pgervais\}@google.com}}

\date{Received: Nov.\ 2018 / Accepted: -}

\maketitle

\begin{abstract}
  We describe an online handwriting system that is able to support 102 languages using a deep neural network architecture. 
  This new system has completely replaced our previous Segment-and-Decode-based system and reduced the error rate by 20\%-40\% relative for most languages. 
  Further, we report new state-of-the-art results on \iamondb for both the open and closed dataset setting.
  The system combines methods from sequence recognition with a new input encoding using \Bezier curves.
  This leads to up to 10x faster recognition times compared to our previous system.
  Through a series of experiments we determine the optimal configuration of our models and report the results of our setup on a number of additional public datasets.
\end{abstract}

\section{Introduction}
\label{sec:introduction}

In this paper we discuss \emph{online} handwriting recognition: Given a user input in the form of an \emph{ink}, i.e. a list of touch or pen \emph{strokes}, output the textual interpretation of this input. A stroke is a sequence of \emph{points} $(x,y,t)$ with position $(x,y)$ and timestamp $t$.

\begin{figure}
  \centering
  \newcommand{\eximg}[2][1.0]{%
    \FPeval\widthresult{#1 * 0.4}
    \includegraphics[width=#1\linewidth,height=\widthresult\linewidth,keepaspectratio]{#2}}
  \begin{minipage}{.3\columnwidth}
    \centering
    \eximg[1]{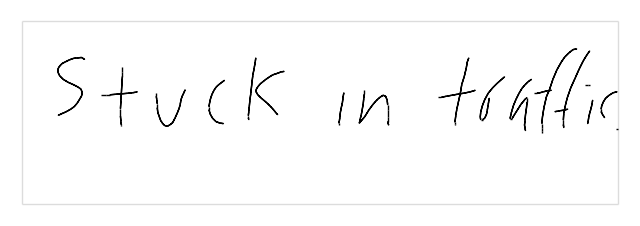}\\
    \eximg[1]{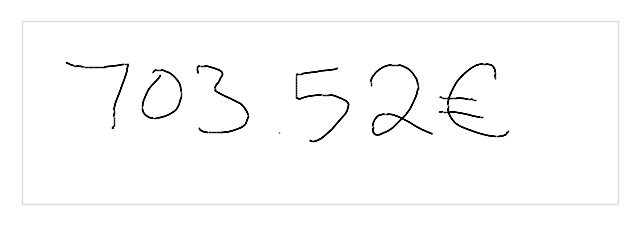}\\
    \eximg[1]{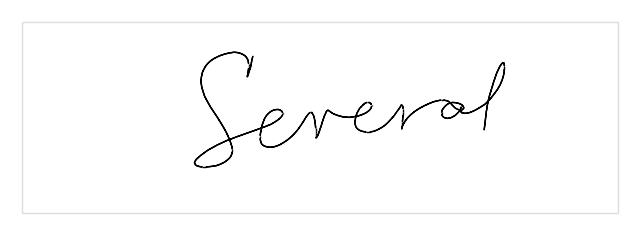}\\
    \eximg[1]{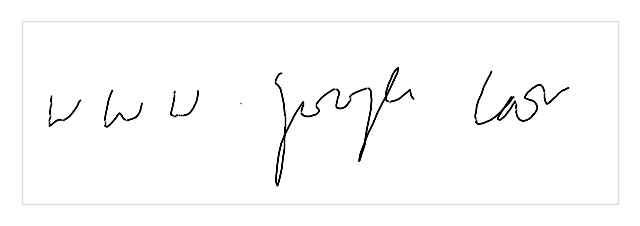}\\
    \eximg[1]{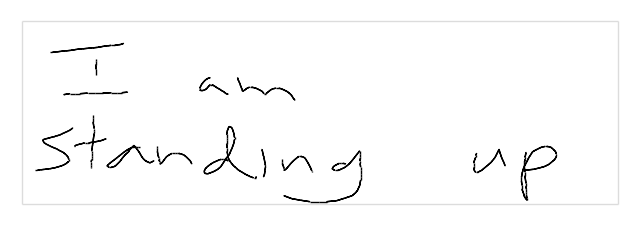}
  \end{minipage}~~~%
  \begin{minipage}{.3\columnwidth}
    \centering
    \eximg[1]{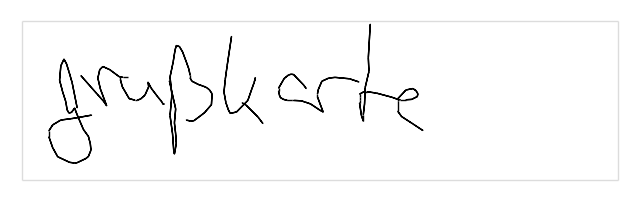}\\
    \eximg[1]{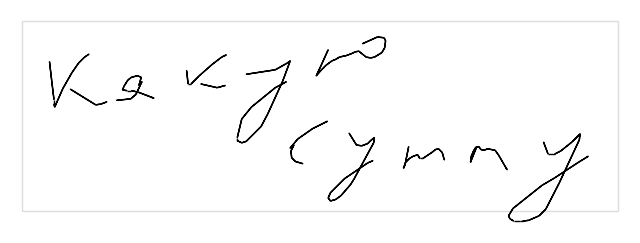}\\
    \eximg[1]{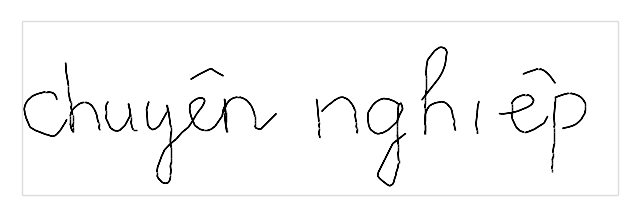}\\
    \eximg[1]{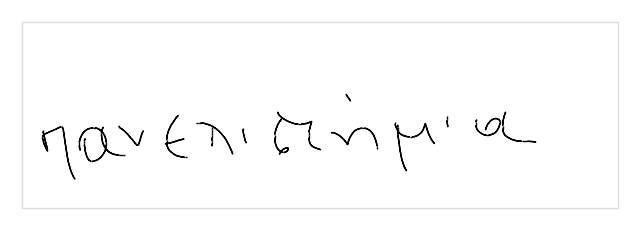}\\
    \eximg[1]{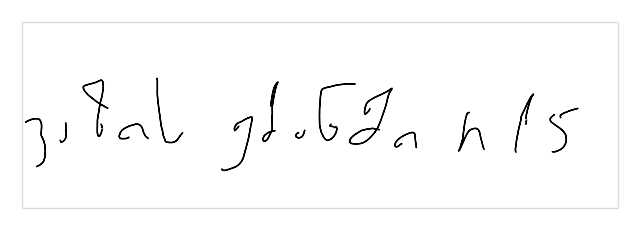}
  \end{minipage}~%
  \begin{minipage}{.3\columnwidth}
    \centering
    \eximg[1]{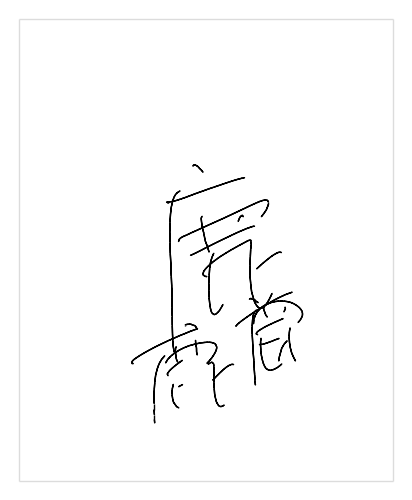}
    \eximg[1]{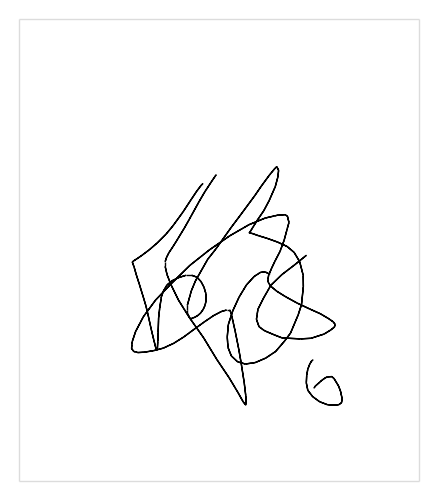}\\
    \eximg[1]{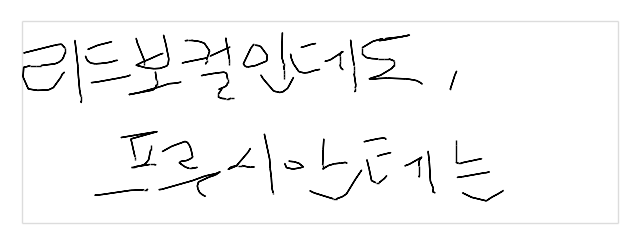}\\
    \eximg[1]{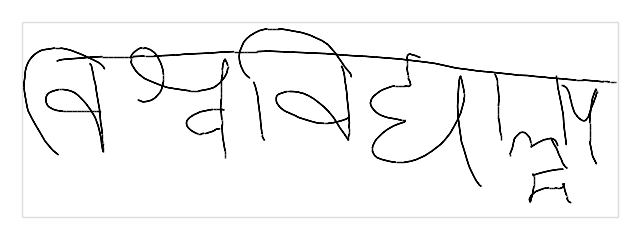}\\
    \eximg[1]{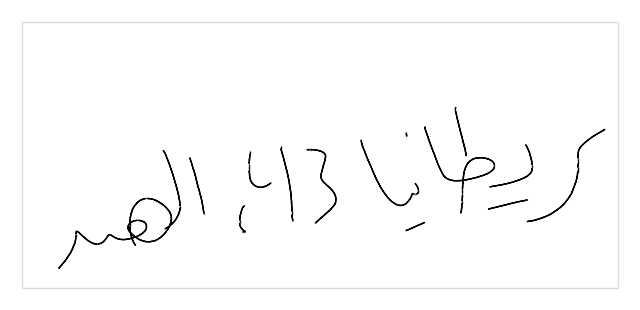}\\
    \eximg[1]{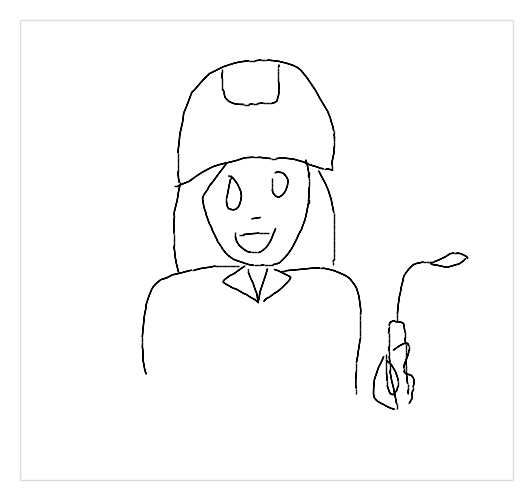}
  \end{minipage}
  \caption{Example inputs for online handwriting recognition in different languages. See text for details.}
\label{fig:example_inputs}
\end{figure}

Figure~\ref{fig:example_inputs} illustrates example inputs to our online handwriting recognition system in different languages and scripts. 
The left column shows examples in English with different writing styles, with different types of content, and that may be written on one or multiple lines. 
The center column shows examples from five different alphabetic languages similar in structure to English: German, Russian, Vietnamese, Greek, and Georgian. 
The right column shows scripts that are significantly different from English:
Chinese has a much larger set of more complex characters, and users often overlap characters with one another. Korean, while an alphabetic language, groups letters in syllables leading to a large ``alphabet'' of syllables. 
Hindi writing often contains a connecting ‘Shirorekha’ line and characters can form larger structures (grapheme clusters) which influence the written shape of the components. 
Arabic is written right-to-left (with embedded left-to-right sequences used for numbers or English names) and characters change shape depending on their position within a word. 
Emoji are non-text Unicode symbols that we also recognize.

Online handwriting recognition has recently been gaining importance for multiple reasons: 
(a) An increasing number of people in emerging markets are obtaining access to computing devices, many exclusively using mobile devices with touchscreens. Many of these users have native languages and scripts that are not as easily typed as English, e.g.\ due to the size of the alphabet or the use of grapheme clusters which makes it difficult to design an intuitive keyboard layout~\cite{ghoshindickeyboardlayout}.
(b) More and more large mobile devices with styluses are becoming available, such as the iPad Pro\footnote{\url{https://www.apple.com/ipad-pro/}}, Microsoft Surface devices\footnote{\url{https://www.microsoft.com/en-us/store/b/surface}}, and Chromebooks with styluses\footnote{\url{https://store.google.com/product/google_pixelbook}}.

Early work in online handwriting recognition looked at segment-and-decode classifiers, such as the Newton \cite{yaeger:aaai98}. Another line of work \cite{plamondon2000online} focused on solving online handwriting recognition  by making use of Hidden Mar\-kov Models (HMMs) \cite{hu1996hmm} or hybrid approaches combining HMMs and Feed-forward Neural Networks \cite{bengio1995lerec}. The first HMM-free models were based on Time Delay Neural Networks (TDNNs) \cite{franzini1990connectionist,jaeger:ijdar01,pittman:2007}, and more recent work focuses on Recurrent Neural Network (RNN) variants such as Long-Short-Term-Memory networks (LSTMs) \cite{graves2008unconstrained,deepblstm-icdar15,frinken:bangla}.

How to represent online handwriting data has been a research topic for a long time. Early approaches were feature-based, where each point is represented using a set of features~\cite{jaeger:ijdar01,jaeger:ijdar03,yaeger:aaai98}, or using global features to represent entire characters~\cite{jaeger:ijdar01}.
More recently, the deep learning revolution has swept away most feature engineering efforts and replaced them with learned representations in many domains, e.g.\  speech~\cite{hinton2012deep}, computer vision~\cite{simonyan2014very}, and natural language processing~\cite{mikolov2013distributed}.

Together with architecture changes, training methodologies also changed, moving from relying on explicit segmentation \cite{pittman:2007,yaeger:aaai98,Google:HWRPAMI} to implicit segmentation using the Connectionist Temporal Classification (CTC) loss \cite{Graves2006ConnectionistTC}, or En\-coder-Decoder approaches trained with Maximum Likelihood Estimation~\cite{DBLP:journals/corr/abs-1712-03991}. Further recent work is also described in \cite{kim:handbook14}.

The transition to more complex network architectures and end-to-end training can be associated with breakthroughs in related fields focused on sequence understanding where deep learning methods have outperformed ``traditional'' pattern recognition methods, e.g.\ in speech recognition \cite{sainath2015deep,sainath2015convolutional}, OCR \cite{Wang2012EndtoendTR,aksara-2017}, offline handwriting recognition  \cite{graves2009offline}, and computer vision \cite{szegedy2016rethinking}.

In this paper we describe our new online handwriting recognition system based on deep learning methods. It replaces our previous segment-and-decode system~\cite{Google:HWRPAMI}, which first over-segments the ink, then groups the segments into character hypotheses, and computes features for each character hypothesis which are then classified as characters using a rather shallow neural network. The recognition result is then obtained using a best path search decoding algorithm on the lattice of hypotheses incorporating additional knowledge sources such as language models. This system relies on numerous pre-processing, segmentation, and feature extraction heuristics which are no longer present in our new system.
The new system reduces the amount of customization required, and consists of a simple stack of bidirectional LSTMs (BLSTMs), a single Logits layer, and the CTC loss \cite{graves:pami09} (Sec.~\ref{sec:model}). We train a separate model for each \textit{script} (Sec.~\ref{sec:training}). 
To support potentially many languages per script (see Table~\ref{tab:script_language}), language-specific language models and feature functions are used during decoding (Sec.~\ref{sec:lms}). E.g.\ we have a single recognition model for Arabic script which is combined with specific language models and feature functions for our Arabic, Persian, and Urdu language recognizers. Table~\ref{tab:script_language} shows the full list of scripts and languages that we currently support.

\begin{table}
    \caption{List of languages supported in our system grouped by script.}
    \label{tab:script_language}
    \centering
    \resizebox{\linewidth}{!}{
    \begin{tabular}{|l|p{0.8\linewidth}|}
    \hline Script & Languages \\ \hline\hline
Latin & Afrikaans, Azerbaijani, Bosnian, Catalan, Cebuano, Corsican, Czech, Welsh, Danish, German, English, Esperanto, Spanish, Estonian, Basque, Finnish, Filipino, French, Western Frisian, Irish, Scottish Gaelic, Galician, Hawaiian, Hmong, Croatian, Haitian Creole, Hungarian, Indonesian, Icelandic, Italian, Javanese, Kurdish, Latin, Luxembourgish, Lao, Lithuanian, Latvian, Malagasy, Maori, Malay, Maltese, Norwegian, Dutch, Nyanja, Polish, Portuguese, Romanian, Slovak, Slovenian, Samoan, Shona, Somali, Albanian, Sundanese, Swedish, Swahili, Turkish, Xhosa, Zulu \\ \hline
Cyrillic & Russian, Belarusian, Bulgarian, Kazakh, Mongolian, Serbian, Ukrainian, Uzbek, Macedonian, Kyrgyz, Tajik \\ \hline
Chinese & Simplified Chinese, Traditional Chinese, Cantonese \\ \hline
Arabic & Arabic, Persian, Urdu \\ \hline
Devanagari & Hindi, Marathi, Nepali \\ \hline
Bengali & Bangla, Assamese \\ \hline
Ethiopic  & Amharic, Tigrinya  \\ \hline
\multicolumn{2}{|p{1.02\linewidth}|}{
Languages with distinct scripts: 
Armenian, 
Burmese, 
Georgian, 
Greek, 
Gujarati, 
Hebrew,
Japanese, 
Kannada, 
Khmer, 
Korean, 
Lao, 
Malayalam, 
Odia, 
Punjabi, 
Sinhala, 
Tamil, 
Telugu, 
Thai, 
Tibetan,
Vietnamese\footnote{While Vietnamese is a Latin script language, we have a dedicated model for it because of the large amount of diacritics not used in other Latin script languages.}
}
\\\hline
    \end{tabular}}
\end{table}

The new models are more accurate (Sec.~\ref{sec:experiments}), smaller, and faster (Table~\ref{tab:english_improvements}) than our previous segment-and-decode models and eliminate the need for a large number of engineered features and heuristics.

We present an extensive comparison of the differences in recognition accuracy for eight languages (Sec.~\ref{sec:performance}) and compare the accuracy of models trained on publicly available datasets where available (Sec.~\ref{sec:experiments}). In addition, we propose a new standard experimental protocol for the \IBMUB data\-set~\cite{IBMUB1} to enable easier comparison between approaches in the future (Sec.~\ref{sec:ibm-ub}).

The main contributions of our paper are as follows:
\begin{itemize}
    \item {We describe in detail our recurrent neural network-based recognition stack and provide a description of how we tuned the model. We also provide a detailed experimental comparison with the previous segment-and-decode based stack \cite{Google:HWRPAMI} on the supported languages.}
    \item {We describe a novel input representation based on \Bezier curve interpolation, which produces shorter input sequences, which results in faster recognitions.}
    \item {Our system achieves a new state-of-the-art on the \iamondb dataset, both for open and closed training sets.}
    \item {We introduce an evaluation protocol for the less commonly-used English \IBMUB query dataset. We provide experimental results that quantify the structural difference between \IBMUB, \iamondb, and our internal dataset.}
    \item {We perform ablation studies and report results on numerous experiments highlighting the con\-tri\-bu\-tions of the individual components of the new recognition stack on our internal datasets.}
\end{itemize}

\section{End-to-end Model Architecture}
\label{sec:model}
\begin{figure*}
\centering
\includegraphics[width=0.7\textwidth]{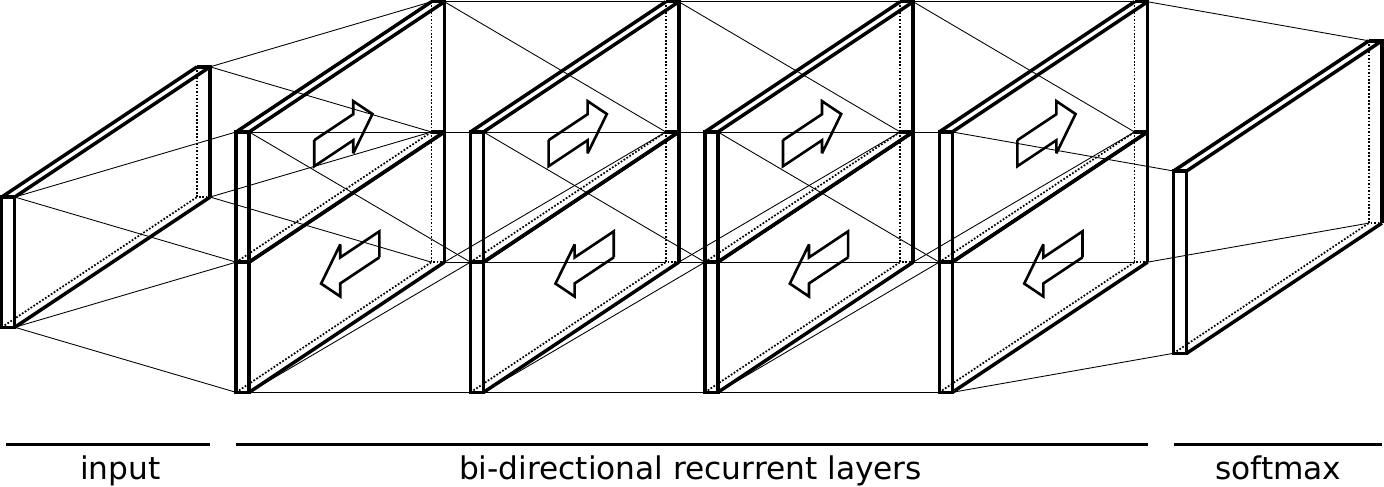}
\label{fig:network}
\caption{An overview our recognition models. In our architecture the input representation is passed through one or more bidirectional LSTM layers, and a final softmax layer makes a classification decision for the output at each time step.}
\end{figure*}

Our handwriting recognition model draws its inspiration from research aimed at building end-to-end transcription models in the context of handwriting recognition \cite{graves:pami09}, optical character recognition \cite{aksara-2017}, and acoustic modeling in speech recognition~\cite{sainath2015deep}. 
The model architecture is constructed from common neural network blocks, i.e.\ bidirectional LSTMs and fully-connected layers (Figure~\ref{fig:network}). It is trained in an end-to-end manner using the CTC loss \cite{graves:pami09}. 

Our architecture is similar to what is often used in the context of acoustic modeling for speech recognition \cite{sainath2015convolutional}, in which it is referred to as a CLDNN (Convolutions, LSTMs, and DNNs), yet we differ from it in four points. 
Firstly, we do not use convolution layers, which in our own experience do not add value for large networks trained on large datasets of relatively short (compared to speech input) sequences  typically seen in handwriting recognition.
Secondly, we use \emph{bidirectional} LSTMs, which due to latency constraints is not feasible in speech recognition systems.
Thirdly, our architecture does not make use of additional fully-connected layers before and after the bidirectional LSTM layers.
And finally, we train our system using the CTC loss, as opposed to the HMMs used in \cite{sainath2015convolutional}.

This structure makes many components of our previous system \cite{Google:HWRPAMI} unnecessary, e.g.\ feature extraction and segmentation. The heuristics that were hard-coded into our previous system, e.g.\ stroke-reordering and character hypothesis building, are now implicitly learned from the training data.

The model takes as input a time series $(v_1,\dots,v_T)$ of length $T$ encoding the user input (Sec.~\ref{sec:processors}) and passes it through several bidirectional LSTM layers~\cite{schuster1997bidirectional} which learn the structure of characters (Sec.~\ref{sec:lstms}).

The output of the final LSTM layer is passed through a softmax layer (Sec.~\ref{sec:softmax}) leading to a sequence of probability distributions over characters for each time step.

For CTC decoding (Sec.~\ref{sec:ctc}) we use beam search to combine the softmax outputs with character-based language models, word-based language models, and information about language-specific characters as in our previous system~\cite{Google:HWRPAMI}.

\subsection{Input Representation}
\label{sec:processors}

In our earlier paper \cite{Google:HWRPAMI} we presented results on our datasets with a model similar to the one proposed in \cite{graves:pami09}. In that model we used 23 per-point features (similarly to \cite{jaeger:ijdar01}) as described in our segment-and-decode system to represent the input. 
In further experimentation we found that in substantially deeper and wider models, engineered features are unnecessary and their removal leads to better results. This confirms the observation that learned representations often outperform handcrafted features in scenarios in which sufficient training data is available, e.g. in computer vision \cite{lecun-lenet98} and in speech recognition \cite{waibel-phonemes89}.
In the experiments presented here, we use two representations:

\subsubsection{Raw Touch Points}
\label{sec:raw}

The simplest representation of stroke data is as a sequence of touch points. In our current system, we use a sequence of 5-dimensional points $(x_i, y_i, t_i, p_i, n_i)$ where $(x_i, y_i)$ are the coordinates of the $i$th touchpoint, $t_i$ is the timestamp of the touchpoint since the first touch point in the current observation in seconds, $p_i$ indicates whe\-ther the point corresponds to a pen-up ($p_i = 0$) or pen-down ($p_i = 1$) stroke, and $n_i = 1$ indicates the start of a new stroke ($n_i = 0$ otherwise).\footnote{We acknowledge a redundancy between the features $p_i$ and $n_i$ which evolved over time from experimenting with explicit pressure data. We did not perform additional experiments to avoid this redundancy at this time but do not expect a large change in results when dropping either of these features.}

In order to keep the system as flexible as possible with respect to differences in the writing surface, e.g. area shape, size, spatial resolution, and sampling rate, we perform some minimal preprocessing:

\begin{itemize}
    \item Normalization of $x_i$ and $y_i$ coordinates, by shifting in $x$ such that $x_0=0$, and shifting and scaling the writing area isometrically such that the $y$ coordinate spans the range between $0$ and $1$. In cases where the bounding box of the writing area is unknown we use a surrogate area 20\% larger than the observed range of touch points. 
    \item Equidistant linear resampling along the strokes with $\delta=0.05$, i.e. a line of length 1 will have 20 points. 
\end{itemize}

\noindent We do not assume that words are written on a fixed baseline or that the input is horizontal.
As  in \cite{graves:pami09}, we use the differences between consecutive points for the $(x,y)$ coordinates and the time $t$ such that our input sequence is $(x_i - x_{i-1}, y_i - y_{i-1}, t_i - t_{i-1}, p_i, n_i)$ for $i>0$, and $(0, 0, 0, p_0, n_0)$ for $i=0$.

\subsubsection{\Bezier Curves}
\label{sec:curves}

However simple, the raw input data has some drawbacks, i.e.
\begin{itemize}
    \item Resolution: Not all input devices sample inputs at the same rate, resulting in different point densities along the input strokes, requiring resampling which may inadvertently normalize-out details in the input.
    \item Length: We choose the (re-)sampling rate such as to represent the smallest features well, which leads to over-sampling in less interesting parts of the stroke, e.g. in straight lines.
    \item Model complexity: The model has to learn to map small consecutive steps to larger global features.
\end{itemize}

\noindent\emph{\Bezier curves} are a natural way to describe trajectories in space, and have been used to represent online handwriting data in the past, yet mostly as a means of removing outliers in the input data~\cite{huang2007preprocessing}, up-sampling sparse data~\cite{jaeger:ijdar01}, or for rendering handwriting data smoothly on a screen~\cite{bezier01}. 
Since a sequence of \Bezier curves can represent a potentially long point sequence compactly, irrespective of the original sampling rate, we propose to represent a sequence of input points as a sequence of parametric cubic polynomials, and to use these as inputs to the recognition model. 

These \Bezier curves for $x$, $y$, and $t$ are cubic polynomials in $s\in[0,1]$:
\begin{equation}
  \label{eq:cubicpoly}
  \begin{array}{rcrcrcrcr}
    x(s) &=& \alpha_{0} &+& \alpha_{1} s &+& \alpha_{2} s^2 &+& \alpha_{3} s^3\\
    y(s) &=& \beta_{0} &+& \beta_{1} s &+& \beta_{2} s^2 &+& \beta_{3} s^3 \\
    t(s) &=& \gamma_{0} &+& \gamma_{1} s &+& \gamma_{2} s^2 &+& \gamma_{3} s^3
  \end{array}
\end{equation}

\noindent We start by normalizing the size of the entire ink such that the $y$ values are within the range $[0, 1]$, similar to how we process it for raw points.
The time values are scaled linearly to match the length of the ink such that
\begin{equation}
    t_{N-1} - t_0 = \sum_{i=1}^{N-1} \left[(x_i - x_{i-1})^2 + (y_i - y_{i-1})^2\right]^{1/2}.
\end{equation}
in order to obtain values in the same numerical range as $x$ and $y$.
This sets the time difference between the first and last point of the stroke to be equal to the total spatial length of the stroke.

For each stroke in an ink, the coefficients $\alpha$, $\beta$, and $\gamma$ are computed by minimizing the sum of squared errors (SSE) between each observed point $i$ and its corresponding closest point (defined by $s_i$) on the \Bezier curve:
\begin{equation}
    \sum_{i=0}^{N-1} \left(x_i - x(s_i)\right)^2 + \left(y_i - y(s_i)\right)^2 + \left(t_i - t(s_i)\right)^2.
    \label{eq:sse}
\end{equation}
\noindent Where $N$ is the number of points in the stroke. Given a set of coordinates $s_i$, computing the coefficients corresponds to solving the following linear system of equations:
\begin{equation}
    \underbrace{\left[\begin{array}{ccc}
        x_1 & y_1 & t_1 \\
        x_2 & y_2 & t_2 \\
        \vdots & \vdots & \vdots \\
        x_N & y_N & t_N
    \end{array}\right]}_{Z} =
    \underbrace{\left[\begin{array}{cccc}
        1 & s_1 & s_1^2 & s_1^3 \\
        1 & s_2 & s_2^2 & s_2^3 \\
        \vdots &\vdots &\vdots &\vdots \\
        1 & s_N & s_N^2 & s_N^3
    \end{array}\right]}_{V}
    \underbrace{\left[\begin{array}{ccc}
        \alpha_0 & \beta_0 & \gamma_0 \\
        \alpha_1 & \beta_1 & \gamma_1 \\
        \alpha_2 & \beta_2 & \gamma_2 \\
        \alpha_3 & \beta_3 & \gamma_3
    \end{array}\right]}_{\Omega}
\end{equation}
\noindent which can be solved exactly for $N\le 4$, and in the least-squares sense otherwise, e.g. by solving the normalized equations
\begin{equation}
    V^\mathsf{T}Z = V^\mathsf{T}V\Omega. \label{coeffs}
\end{equation}
\noindent for the coefficients $\Omega$.
We alternate between minimizing the SSE in eq.~(\ref{eq:sse}) and finding the corresponding points $s_i$, until convergence. 
The coordinates $s_i$ are updated using a Newton step on
\begin{equation}
    x'(s_i)(x_i - x(s_i)) + y'(s_i)(y_i - y(s_i)) = 0,
    \label{coord}
\end{equation}
\noindent which is zero when $(x_i-x(s_i), y_i-y(s_i))$ is orthogonal to the direction of the curve $(x'(s_i), y'(s_i))$.

If (a) the curve cannot fit the points well (SSE error is too large) or if (b) the curve has too sharp bends (arc length longer than 3 times the endpoint distance) we split the curve into two parts.
We determine the split point in case (a) by finding the triplet of consecutive points with the smallest angle, and in case (b) as the point closest to the maximum local curvature along the entire \Bezier curve. This heuristic is applied recursively until both the curve matching criteria are met.

As a final step, to remove spurious breakpoints, consecutive curves that can be represented by a single curve are stitched back together, resulting in a compact set of \Bezier curves representing the data within the above constraints. For each consecutive pair of curves, we try to fit a single curve using the combined set of underlying points. If the fit agrees with the above criteria, we replace the two curves by the new one. This is applied repeatedly until no merging happens anymore.

\begin{figure}
    \centering
    \includegraphics[width=0.6\linewidth]{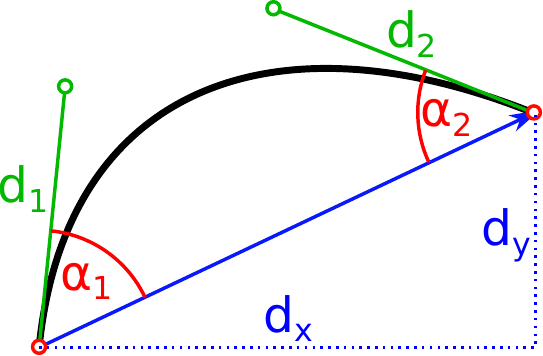}
    \caption{Parameterization of each \Bezier curve used to feed the network. Namely: vector between the endpoints (blue), distance between the control points and the endpoints (green dashed lines, 2 values), and the two angles between each control point and the endpoints (green arcs, 2 values).}
    \label{fig:bezier}
\end{figure}

Since the \Bezier coefficients $\alpha$, $\beta$, and $\gamma$ may vary significantly in range, each curve is fed to the network as a 10-dimensional vector $(d_x, d_y, d_1, d_2, \alpha_1, \alpha_2, \gamma_1, \gamma_2, \gamma_3, p)$, with:
\begin{itemize}
    \item $d_x, d_y$: the vector between the endpoints (cp.\ figure~\ref{fig:bezier})
    \item $d_1, d_2$: the distance between the control points and the endpoints relative to the distance between the endpoints (cp.\ figure~\ref{fig:bezier}),
    \item $\alpha_1, \alpha_2$: the angles between control points and endpoints in radians (cp.\ figure~\ref{fig:bezier}),
    \item $\gamma_1$, $\gamma_2$ and $\gamma_3$: the time coefficients from eq.~\ref{eq:cubicpoly},
    \item $p$: a Boolean value indicating whether this is a pen-up or pen-down curve.
\end{itemize}
\noindent Due to the normalization of the $x$, $y$, and $t$ coordinates, as well as the constraints on the curves themselves, most of the resulting values are in the range $[-1,1]$.

The resulting sequences of 10-dimensional curve representations are roughly 4$\times$ shorter than the corresponding 5-dimensional raw representation (sec.\ref{sec:raw}) because each Bezier curve represents multiple points. This leads to faster recognition and thus better latency.

In most of the cases, as highlighted through the experimental sections in this paper, the curve representations do not have a big impact on accuracy but contribute to faster speed of our models.

\subsection{Bidirectional Long-Short-Term-Memory Recurrent Neural Networks}
\label{sec:lstms}

LSTMs \cite{Hochreiter1997LSTM} have become one of the most commonly used RNN cells because they are easy to train and give good results \cite{jozefowicz2015empirical}. 
In all experiments we use bidirectional LSTMs \cite{Graves2006ConnectionistTC,frinken:bangla}, i.e. we process the input sequence forward and backward and merge the output states of each layer before feeding them to the next layer. 
The exact number of layers and nodes is determined empirically for each script.
We give an overview of the impact of the number of nodes and layers in section~\ref{sec:experiments}. We also list the configurations for several scripts in our production system, as of this writing.

\subsection{Softmax Layer}
\label{sec:softmax}

The output of the LSTM layers at each timestep is fed into a softmax layer to get a probability distribution over the $C$ possible characters in the script (including spaces, punctuation marks, numbers or other special characters), plus the blank label required by the CTC loss and decoder.

\subsection{Decoding}
\label{sec:decoding}

The output of the softmax layer is a sequence of $T$ time steps of $(C+1)$ classes  that we decode using CTC decoding \cite{Graves2006ConnectionistTC}.
The logits from the softmax layer are combined with language-specific prior knowledge (cp. Sec.~\ref{sec:lms}).
For each of these additional knowledge sources we learn a weight (called ``decoder weight'' in the following) and combine them linearly (cp. Sec.~\ref{sec:training}).
The learned combination is used as described in \cite{Graves2014TowardsES} to guide the beam search during decoding.\footnote{We implement this as a {\tt BaseBeamScorer} (\url{https://github.com/tensorflow/tensorflow/blob/master/tensorflow/core/util/ctc/ctc_beam_scorer.h}) which is passed to the {\tt CTCBeamSearchDecoder} implementation in TensorFlow \cite{tensorflow2015-whitepaper}.} 

This combination of different knowledge sources allows us to train one recognition model per script (e.g.\ Latin script, or Cyrillic script) and then use it to serve multiple languages (see Table~\ref{tab:script_language}).

\subsection{Feature Functions: Language Models and Character Classes}
\label{sec:lms}

Similarly to our previous work \cite{Google:HWRPAMI}, we define several scoring functions, which we refer to as \emph{feature functions}. The goal of these feature functions is to introduce prior knowledge about the underlying language into the system. The introduction of recurrent neural networks has reduced the need for many of them and we now use only the following three:

\begin{itemize}
    \item Character Language Models:
    For each language we support, we build a 7-gram language model over Unicode codepoints from a large web-mined text corpus using Stupid back-off~\cite{brants:prodlm2007}. The final files are pruned to \num{10} million \num{7}-grams each. 
    Compared to our previous system~\cite{Google:HWRPAMI}, we found that language model size has a smaller impact on the recognition accuracy, which is likely due to the capability of recurrent neural networks to capture dependencies between consecutive characters. We therefore use smaller language models over shorter contexts.
    
    \item Word Language Models:
    For languages using spaces to separate words, we also use a word-based language model trained on a similar corpus as the character language models~\cite{speech-lm-mining,speech-lm-infra}, using 3-grams pruned to between \num{1.25} million and \num{1.5} million entries.
    
    \item Character Classes:
    We add a scoring heuristic which boosts the score of characters from the language's alphabet. 
    This feature function provides a strong signal for rare characters that may not be recognized confidently by the LSTM, and which the other language models might not weigh heavily enough to be recognized.
    This feature function was inspired by our previous system \cite{Google:HWRPAMI}.
\end{itemize}

\noindent In Section~\ref{sec:experiments} we provide an experimental evaluation of how much each of these feature functions contributes to the final result for several languages.

\section{Training}
\label{sec:training}

The training of our system happens in two stages, on two different datasets:
\begin{enumerate}
    \item End-to-end training of the neural network model using the CTC loss using a large training dataset
    \item Tuning of the decoder weights using Bayesian optimization through Gaussian Processes in Vizier \cite{golovin-vizier}, using a much smaller and distinct dataset.
\end{enumerate}

\noindent Using separate datasets is important because the neural network learns the local appearance as well as an implicit language model from the training data. It will be overconfident on its training data and thus learning the decoder weights on the same dataset could result in weights biased towards the neural network model.

\subsection{Connectionist Temporal Classification Loss}
\label{sec:ctc}

As our training data does not contain frame-aligned labels, we rely on the CTC loss \cite{Graves2006ConnectionistTC} for training which treats the alignment between inputs and labels as a hidden variable. CTC training introduces an additional blank label  which is used internally for learning alignments jointly with character hypotheses, as described in \cite{Graves2006ConnectionistTC}. 

We train all neural network weights jointly using the standard TensorFlow~\cite{tensorflow2015-whitepaper} implementation of CTC training using the Adam Optimizer~\cite{Kingma2014AdamAM} with a batch size of 8, a learning rate of $10^{-4}$, and gradient clipping such that the gradient $L_2$-norm is $\le 9$.
Additionally, to improve the robustness of our models and prevent overfitting, we train our models using random dropout \cite{hinton12dropout,pham2014dropout} after each LSTM layer with a dropout rate of $0.5$.
We train until the error rate on the evaluation dataset no longer improves for 5 million steps. %

\subsection{Bayesian Optimization for Tuning Decoder Weights}
\label{sec:vizier}

To optimize the decoder weights, we rely on the Google Vizier service and its default algorithm, 
specifically batched Gaussian process bandits, and expected improvement as the acquisition function \cite{golovin-vizier}. 

For each recognizer training we start 7 Vizier studies, each performing 500 individual trials, and then we pick the configuration that performed best across all of these trials. We experimentally found that using 7 separate studies with different random initializations regularly leads to better results than running a single study once. We found that using more than 500 trials per study does not lead to any additional improvement.

For each script we train these weights on a subset of the languages for which we have sufficient data, and transfer the weights to all the other languages. E.g.\ for the Latin-script languages, we train the decoder weights on English and German, and use the resulting weights for all languages in the first row of Table~\ref{tab:script_language}.

\section{Experimental Evaluation}
\label{sec:experiments}

In the following, where possible, we present results for public datasets in a \emph{closed data} scenario, i.e. training and testing models on the public dataset using a standard protocol. 
In addition we present evaluation results for public datasets in an \emph{open data} scenario against our production setup, i.e. in which the model is trained on our own data. Finally, we show experimental results for some of the major languages on our internal datasets.  
Whenever possible we compare these results to the state of the art and to our previous system~\cite{Google:HWRPAMI}.

\subsection{\iamondb}
The \iamondb dataset \cite{liwicki:icdar05} is probably the most used evaluation dataset for online handwriting recognition. It consists of 298\,523 characters in 86\,272 word instances from a dictionary of 11\,059 words written by 221 writers. We use the standard \iamondb dataset separation: one training set, two validations sets and a test set containing 5\,363, 1\,438, 1\,518 and 3\,859 written lines, respectively.
We tune the decoder weights using the validation set with 1\,438 items and report error rates on the test set.

\begin{table}
    \caption{Comparison of character error rates (lower is better) on the \iamondb test set for different LSTM layers configurations. For each LSTM width and input type, we show   the best result in bold.}
    \label{tab:iamondb_raw_vs_curves}
    \centering
    \newcommand{\w}{\multicolumn{1}{c|}{yes}}
    \newcommand{\wo}{\multicolumn{1}{|c||}{no}}
    \resizebox{\linewidth}{!}{
    \begin{tabular}{|l|l||r|r|r|}
    \hline
      input & lstm & 64 nodes & 128 nodes & 256 nodes 
      \\\hline\hline
      &1 layer  & 6.1      & 5.95     & 5.56     \\
raw   &3 layers & \bf 4.03 & 4.73     & 4.34     \\
      &5 layers & 4.34     & \bf 4.20 & \bf 4.17 \\
    \hline
    \hline
       &1 layer  & 6.57     & 6.38     &  6.98    \\
curves &3 layers & 4.16     & \bf 4.16 & 4.83     \\
       &5 layers & \bf 4.02 & 4.22     & \bf 4.11 \\
\hline
    \end{tabular}}

\end{table}

\begin{figure*}
\includegraphics[width=.49\linewidth]{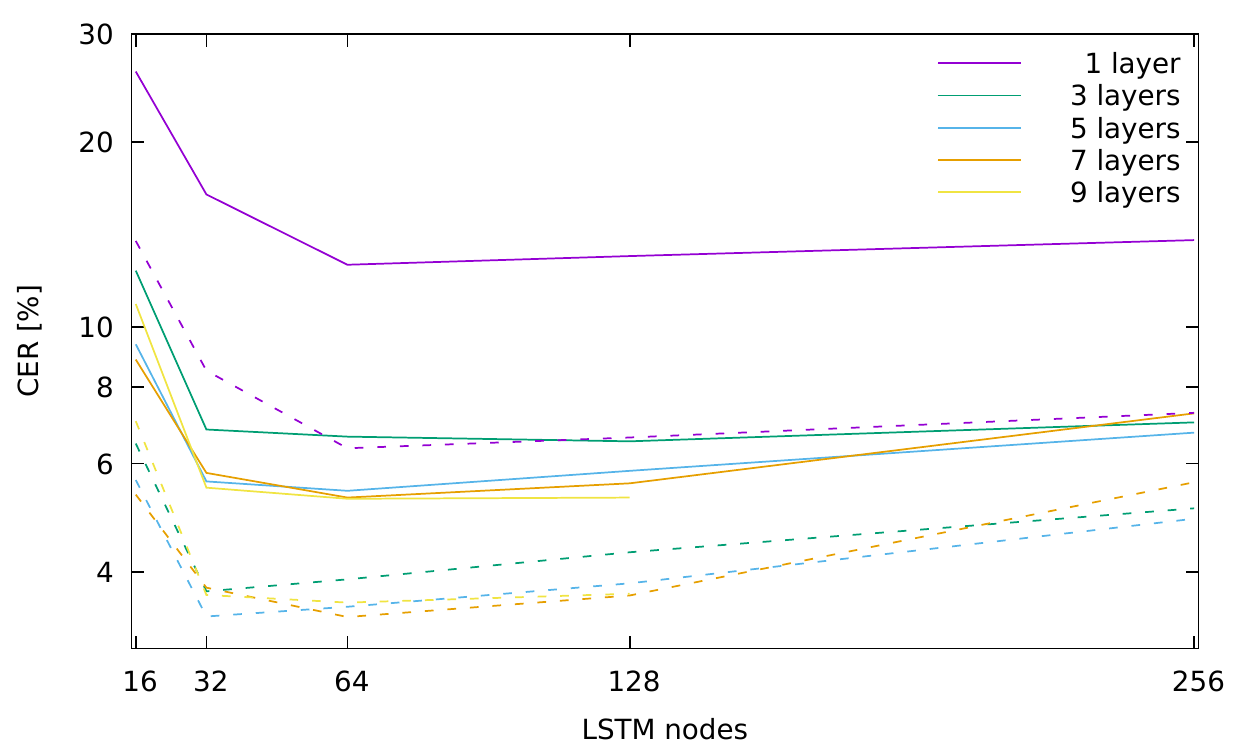}
\includegraphics[width=.49\linewidth]{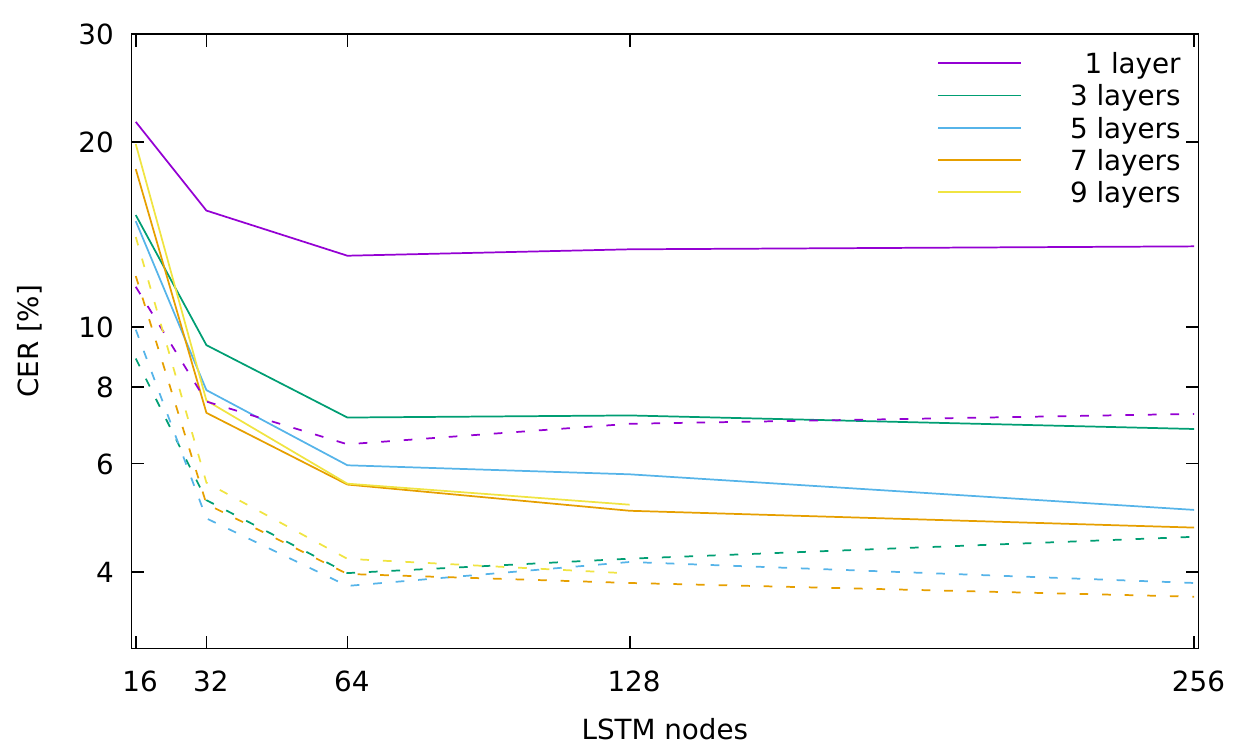}
\caption{CER of models trained on the \iamondb dataset with different numbers of LSTM layers and LSTM nodes using raw (left) and curve (right) inputs. Solid lines indicate results without any language models or feature functions in decoding, dashed lines indicate results with the fully-tuned system.}
\label{fig:iamondb_lstms_stuff}
\end{figure*}

We perform a more extensive study of the number of layers and nodes per layer for both the raw and curve input formats to determine the optimal size of the bidirectional LSTM network (see Figure~\ref{fig:iamondb_lstms_stuff}, Table~\ref{tab:iamondb_raw_vs_curves}). 
We first run experiments without additional feature functions (Figure~\ref{fig:iamondb_lstms_stuff}, solid lines), then re-compute the results with tuned weights for language models and character classes (Figure~\ref{fig:iamondb_lstms_stuff}, dashed lines).
We observe that for both input formats, using 3 or 5 layers outperforms more shallow networks, and using more layers gives hardly any improvement. 
Furthermore, using 64 nodes per layer is sufficient, as wider networks give only small improvements, if at all. We see no significant difference in accuracy between the raw and the curve representation.

\begin{table}
  \centering
  \caption{Error rates on the \iamondb test set in comparison to the state of the art and our previous system \cite{Google:HWRPAMI}.
  A "*" in the "system" column indicates the use of an open training set. "FF" stands for "feature functions" as described in sec.\ \ref{sec:decoding}.}
  \label{tab:iamdb-res}
\newcommand{\rast}{*}
\newcommand{\past}{\phantom{*}}
  \begin{tabular}{rrr}
                        system                              & CER[\%] & WER[\%] \\
\hline
             Frinken et al.\ BLSTM \cite{deepblstm-icdar15}            \past &  12.3   & 25.0 \\
             Graves et al.\ BLSTM \cite{graves:pami09}                \past &  11.5   & 20.3 \\
             Liwicki et al.\ LSTM \cite{liwicki:mva2011}               \past &    -    & 18.9 \\
             this work   (curve, 5x64, no FF)         \past &   5.9   & 18.6 \\
             this work   (curve, 5x64, FF)          \past &   \bf 4.0   & \bf 10.6 \\ 
\hline
              our previous BLSTM \cite{Google:HWRPAMI} \rast &   8.8   & 26.7 \\
          combination \cite{liwicki:mva2011}           \rast &    -    & 13.8 \\
           our Segment-and-Decode \cite{Google:HWRPAMI}    \rast &   4.3   & 10.4 \\
             this work (production system)             \rast &   \bf 2.5   & \bf 6.5\\
\hline
\end{tabular}
\end{table}

Finally, we show a comparison of our old and new systems with the literature on the \iamondb dataset in Table~\ref{tab:iamdb-res}. Our method establishes a new state of the art result when relying on closed data using \iamondb, as well
as when relying on our in-house data that we use for our production system, which was not tuned for the \iamondb data and for which none of the \iamondb data was used for training. 

To better understand where the improvements come from, we discuss the differences between the previous state-of-the-art system (Graves et al.\ BLSTM \cite{graves:pami09}) and this work across four dimensions: input pre-processing and feature extraction, neural network architecture, CTC training and decoding, and model training methodology.

Our input pre-processing (Sec~\ref{sec:processors}) differs only in minor ways: the $x$-coordinate used is not first transformed using a high-pass filter, we don't split text-lines using gaps and we don't remove delayed strokes, nor do we do any skew and slant correction or other pre-processing.

The major difference comes from feature extraction. In  contrast to the 25 features per point used in \cite{graves:pami09}, we use either 5 features (raw) or 10 features (curves). While the 25 features included both temporal (position in the time series) and spatial features (offline representation), our work uses only the temporal structure. In contrast also to our previous system \cite{Google:HWRPAMI}, using a more compact representation (and reducing the number of points for curves) allows a feature representation, including spatial structure, to be learned in the first or upper layers of the neural network.

The neural network architecture differs both in internal structure of the LSTM cell as well as in the architecture configuration.
Our internal structure differs only in that we do not use peephole connections \cite{gers2001lstm}.

As opposed to relying on a single bidirectional LSTM layer of width 100, we experiment with a number of configuration variants as detailed in Figure~\ref{fig:iamondb_lstms_stuff}. We note that it is particularly important to have more than one layer in order to learn a meaningful representation without feature extraction.

We use the CTC forward-backward training algorithm as described in \cite{graves:pami09}, and implemented in TensorFlow. The training hyperparameters are described in Section~\ref{sec:ctc}.

The CTC decoding algorithm incorporates feature functions similarly to how the dictionary is incorporated in the previous state-of-the-art system.
However, we use more feature functions, our language models are trained on a different corpus, and the combination weights are optimized separately as described in Sec~\ref{sec:vizier}.

\subsection{\IBMUB}
\label{sec:ibm-ub}

Another publicly-accessible English-language dataset is the \IBMUB dataset~\cite{IBMUB1}. From the available datasets therein, we use the English query dataset, which consists of 63\,268 handwritten English words. As this dataset has not been used often in the academic literature, we propose an evaluation protocol. We split this dataset into 4 parts with non-overlapping writer IDs: 47\,108 items for training, 4\,690 for decoder weight tuning, 6\,134 for validation and
5\,336 for testing%
\footnote{Information about the exact experimental protocol is available at \url{https://arxiv.org/src/1902.10525v1/anc}}.

\begin{figure*}
\includegraphics[width=.49\linewidth]{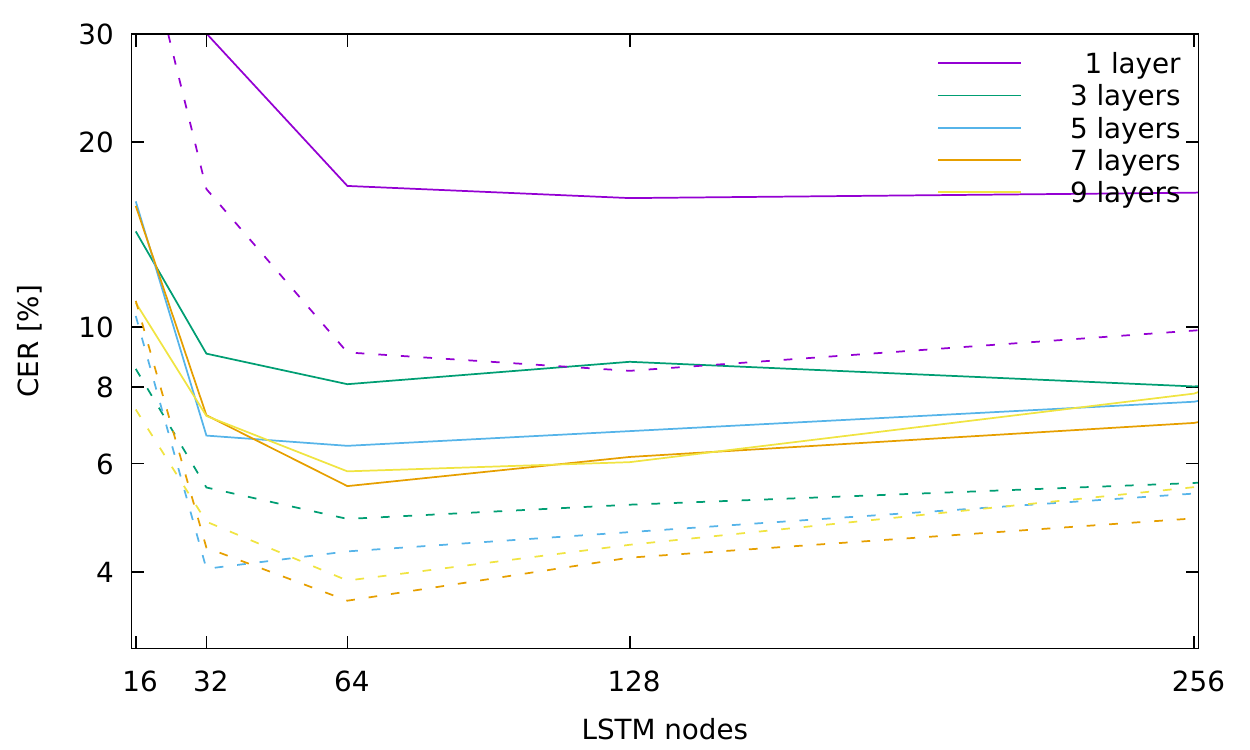}
\includegraphics[width=.49\linewidth]{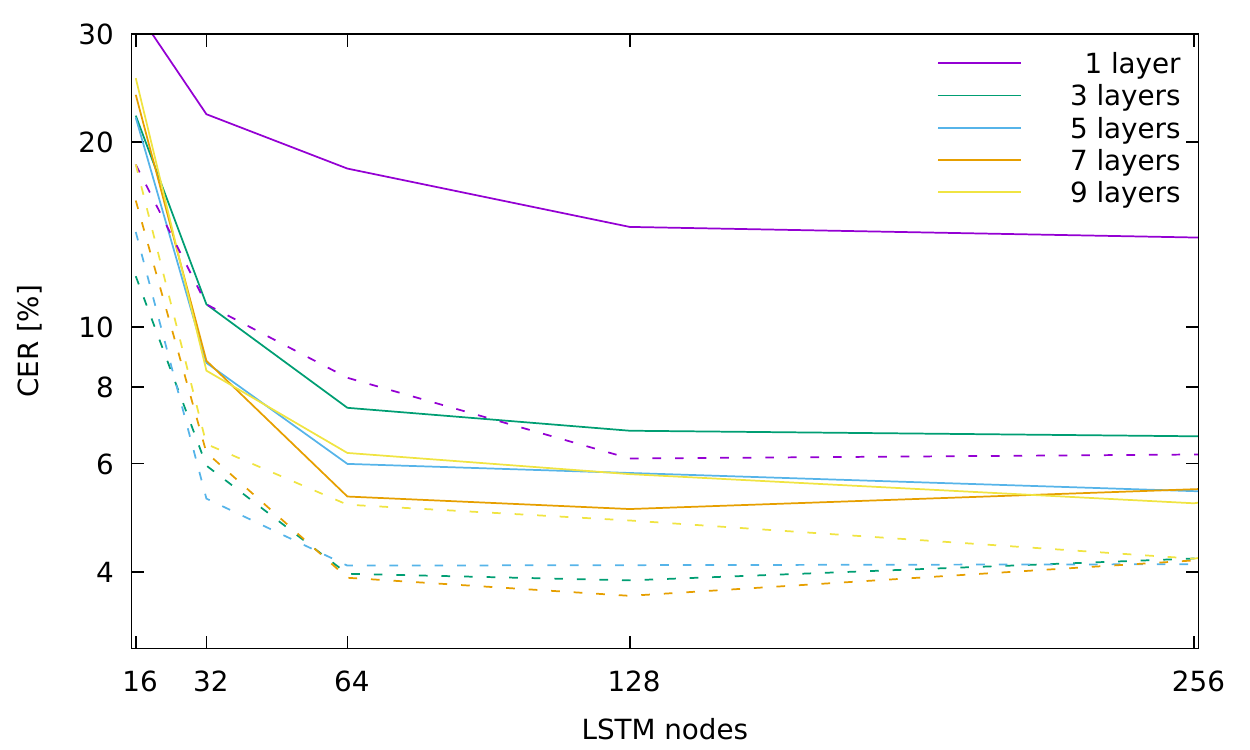}
\caption{CER of models trained on the \IBMUB dataset with different numbers of LSTM layers and LSTM nodes using raw (left) and curve (right) inputs. Solid lines indicate results without any language models or feature functions in decoding, dashed lines indicate results with the fully-tuned system.}
\label{fig:ibmub_lstms_stuff}
\end{figure*}

We perform a similar set of experiments as we did for \iamondb to determine the right depth and width of our neural network architecture.
The results of these experiments are shown in Figure~\ref{fig:ibmub_lstms_stuff}. The conclusion for this dataset is similar to the conclusions we drew for the \iamondb: using networks with 5 layers of bidirectional LSTMs with 64 cells each is sufficient for good accuracy. Less deep and less wide networks perform substantially worse, but larger networks only give small improvements. This is true regardless of the input processing method chosen and again, we do not see a significant difference in accuracy between the raw and curve representation in accuracy.

\begin{table}
  \centering
  \caption{Error rates on \IBMUB test set in comparison to our previous system \cite{Google:HWRPAMI}.
  A "*" in the "system" column indicates the use of an open training set.}
  \label{tab:ibmub-res}
\newcommand{\rast}{*}
\newcommand{\past}{\phantom{*}}
  \begin{tabular}{rrr}
                        system                                      & CER[\%] & WER[\%] \\
\hline
             this work (curve, 5x64, no FF)                        \past &   6.0  &  25.1 \\
             this work (curve, 5x64, FF)                         \past &   4.1  &  15.1 \\
             Segment-and-Decode from \cite{Google:HWRPAMI}          \rast &   6.7  &  22.2 \\
             this work (production system) (sec~\ref{sec:performance})    \rast &   4.1  &  15.3 \\
\hline
\end{tabular}
\end{table}

We give some exemplary results and a comparison with our current production system as well as results for our previous system in Table~\ref{tab:ibmub-res}. We note that our current system is about 38\% and 32\% better (relative) in CER and WER, respectively, when compared to the previous segment-and-decode approach.
The lack of improvement in error rate when evaluating on our production system is due to the fact that our datasets contain spaces while the same setup trained solely on \IBMUB does not.

\subsection{Additional public datasets}

We provide an evaluation of our production system trained on our in-house datasets applied to a number of publicly available benchmark datasets from the literature. More detail about our in-house datasets is available from table~\ref{tab:english_improvements}.
Note that for all experiments presented in this section we evaluate our current live system without any tuning specific to the tasks at hand.

\subsubsection{Chinese Isolated Characters (ICDAR 2013 competition)}

The ICDAR-2013 Competition for Online Handwriting Chinese Character Recognition \cite{yin2013icdar} introduced
a dataset for classifying the most common Chinese characters. We report the error rates in comparison to published results from the competition and more recent work done by others in Table~\ref{tab:icdar2013-chinese}.

We evaluate our live production system on this dataset. Our system was not tuned to the task at hand and was trained as a multi-character recognizer, thus it is not even aware that each sample only contains a single character. Further, our system supports 12\,363 different characters while the competition data only contains 3\,755  characters. Note that our system did not have access to the training data for this task at all.

Whenever our system returns more than one character for a sample, we count this as an error (this happened twice on the entire test set of \num{224\,590} samples). 
Despite supporting almost four times as many characters than needed for the CASIA data and not having been tuned to the task, the accuracy of our system is still competitive with systems that were tuned for this data specifically.

\begin{table}
  \centering
  \caption{Error rates on ICDAR-2013 Competition Database of Online Handwritten Chinese Character Recognition.
  Our system was trained with an open training set, including a mix of characters, words, and phrases.}
  \label{tab:icdar2013-chinese}
\newcommand{\rast}{*}
\newcommand{\past}{\phantom{*}}
  \begin{tabular}{lr}
                        system                                                  & ER[\%] \\
\hline
             Human Performance \cite{yin2013icdar}                              & 4.8 \\
             Traditional Benchmark \cite{liu2013online}                         & 4.7 \\
             ICDAR-2011 Winner \cite{icdar2011-casia-competition}               & 4.2 \\ \hline
             this work (production system) sec.~\ref{sec:performance}           & 3.2 \\ \hline
             ICDAR-2013 Winner: UWarwick \cite{yin2013icdar}                    & 2.6 \\
             RNN: NET4 \cite{casiaRNNnet4}                                      & 2.2 \\
             100LSTM-512LSTM-512FC-3755FC \cite{CASIA-icfhr18}                  & 2.2 \\
             RNN: Ensemble-NET123456 \cite{casiaRNNnet4}                        & 1.9 \\
\hline
\end{tabular}
\end{table}

\subsubsection{Vietnamese Online Handwriting Recognition (ICFHR 2018 competition)}

In the ICFHR2018 Competition on  Vietnamese Online Handwritten Text Recognition using VNOnDB
\cite{viet-comp-icfhr2018}, our production system was evaluated against other systems.
The system used in the competition is the one reported and described in this paper. Due to licensing restrictions we
were unable to do any experiments on the competition training data, or specific tuning for the competition, which was not the case for the other systems mentioned here.

We participated in the two tasks that best suited the purpose of our system,
specifically the "Word" (ref. table~\ref{tab:vnondb2018-word}) and the "Text line" (ref. table~\ref{tab:vnondb2018-line})
recognition levels. Even though we can technically process paragraph level inputs, our system
was not built  with this goal in mind.

\begin{table}
  \centering
  \caption{Results on the VNONDB-Word dataset.}
  \label{tab:vnondb2018-word}
  \resizebox{\columnwidth}{!}{  
      \begin{tabular}{c|c|c|c|c}
         & \multicolumn{2}{c|}{public test set} & \multicolumn{2}{|c}{secret test set} \\
        system & CER[\%] & WER[\%] & CER[\%] & WER[\%] \\
        \hline
        this work (sec.~\ref{sec:performance})    & 6.1 & 13.2 & 9.8 & 20.5 \\
        IVTOVTask1                                & 2.9 &  6.5 & 7.3 & 15.3 \\
        MyScriptTask1                             & 2.9 &  6.5 & 6.0 & 12.7 \\
        \hline
      \end{tabular}
  }
\end{table}

\begin{table}
  \centering
  \caption{Results on the VNONDB-Line dataset.}
  \label{tab:vnondb2018-line}
  \resizebox{\columnwidth}{!}{  
      \begin{tabular}{c|c|c|c|c}
         & \multicolumn{2}{c|}{public test set} & \multicolumn{2}{|c}{secret test set} \\
        system & CER[\%] & WER[\%] & CER[\%] & WER[\%] \\
        \hline
        this work (sec.~\ref{sec:performance})        & 6.9 & 19.0 & 10.3 & 27.0 \\
        IVTOVTask2                                    & 3.2 & 14.1 &  5.6 & 21.0 \\
        MyScriptTask2\_1                              & 1.0 &  2.0 &  1.0 &  3.4 \\
        MyScriptTask2\_2                              & 1.6 &  4.0 &  1.7 &  5.1 \\
        \hline
      \end{tabular}
  }
\end{table}

In contrast to us, the other teams used the training and validation sets to tune their systems:

\begin{itemize}
\item The IVTOV team's system is very similar to our system. It makes use of bidirectional
LSTM layers trained end-to-end with the CTC loss. The inputs used are delta $x$ and $y$ coordinates,
together with pen-up strokes (boolean feature quantifying whether a stroke has ended or not). They report using a
two-layer network of 100 cells each and additional preprocessing for better handling the
dataset.

\item The MyScript team submitted two systems. The first system has an explicit segmentation
component along with a feed-forward network for recognizing character hypotheses, similar
in formulation to our previous system \cite{Google:HWRPAMI}. In addition, they also make
use of a bidirectional LSTM system trained end-to-end with the CTC loss. They do not provide
additional details on which system is which. 
\end{itemize}
We note that the modeling stacks of the systems out-performing ours in this competition are not fundamentally different (to the best of our knowledge, according
to released descriptions). We therefore believe that our system might perform comparably if trained
on the competition training dataset as well.

On our internal testset of Vietnamese data, our new system obtains a CER of 3.3\% which is 54\% relative better than the old Segment-and-Decode system which had a CER of 7.2\% (see also Table~\ref{fig:scatter}).

\subsection{Tuning neural network parameters on our internal data}

Our in-house datasets consist of various types of training data, the amount of which varies by script. Sources of training data include data collected through prompting, commercially available data, artificially inflated data, and labeled/self-labeled anonymized recognition requests (see \cite{Google:HWRPAMI} for a more detailed description).
This leads to more heterogeneous datasets than academic datasets such as \IBMUB\  or \iamondb\  which were collected under standardized conditions.
The number of training samples varies from tens of thousands to several million per script, depending on the complexity and usage. We provide more information about the size of our internal training and tests datasets in table~\ref{tab:english_improvements}.

\begin{figure*}
\includegraphics[width=.49\linewidth]{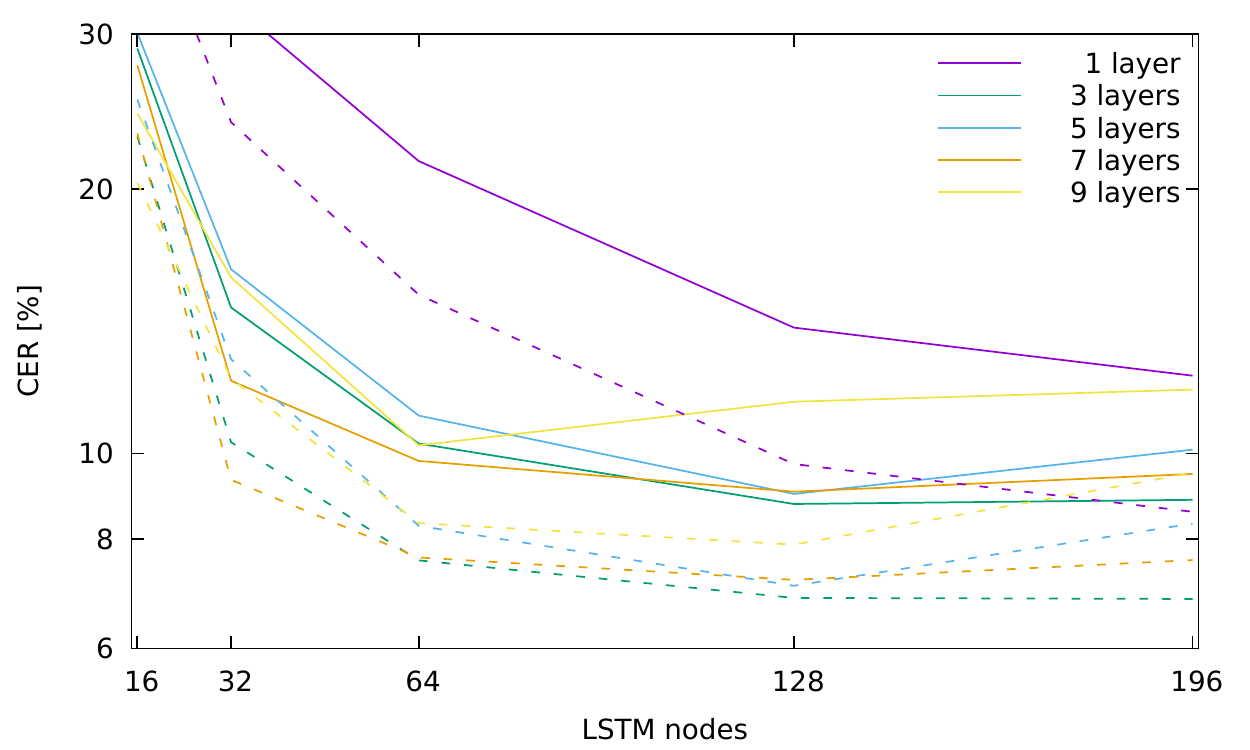}
\includegraphics[width=.49\linewidth]{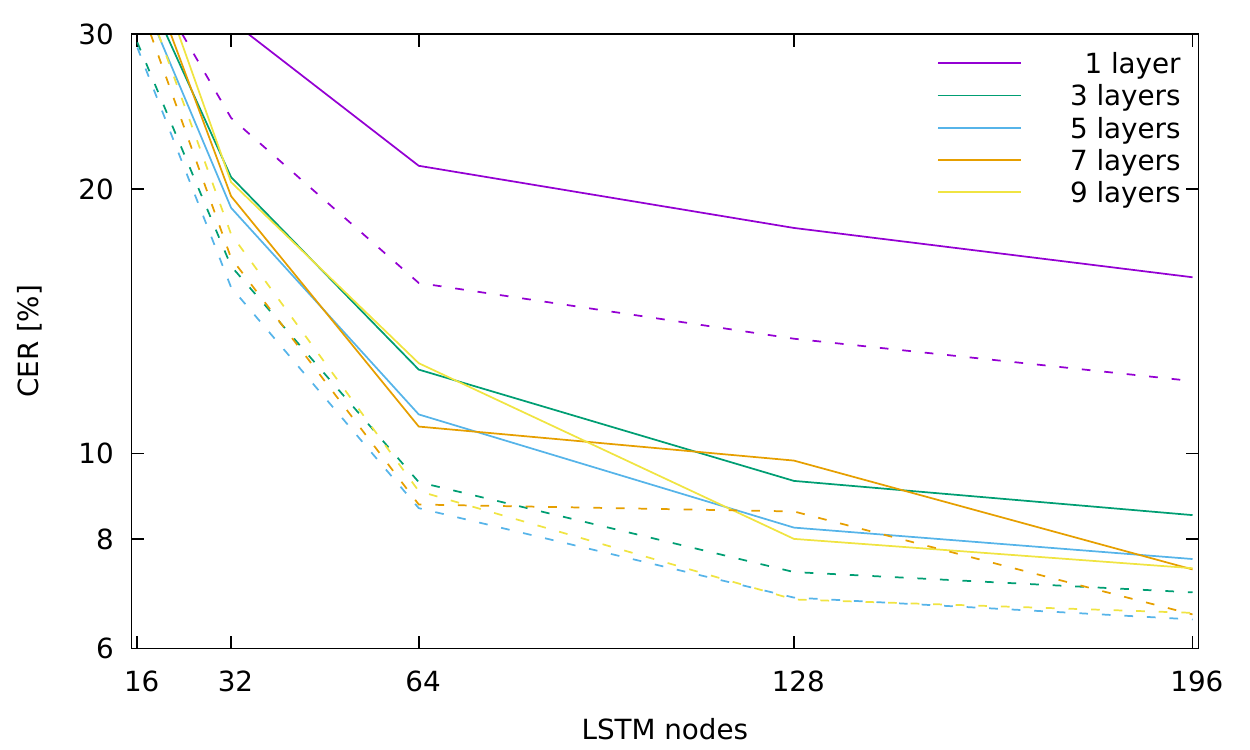}
\caption{CER of models trained on our internal datasets evaluated on our English-language validation set with different numbers of LSTM layers and LSTM nodes using raw (left) and curve (right) inputs. Solid lines indicate results without any language models or feature functions in decoding, dashed lines indicate results with the fully-tuned system.}
\label{fig:latin_en_lstms_stuff}
\end{figure*}

The best configuration for our system was identified by running multiple experiments over a range of layer depths and widths on our internal datasets. For the Latin script experiments shown in Figure~\ref{fig:latin_en_lstms_stuff}, the training set we used was a mixture of data from all the Latin-script languages we support and evaluation is done on an English validation dataset, also used for the English evaluation in Table~\ref{tab:english_improvements}.

Similarly to experiments depicted in Figures~\ref{fig:iamondb_lstms_stuff} and \ref{fig:ibmub_lstms_stuff}, increasing the depth and width of the network architecture brings diminishing returns fairly quickly. However, overfitting is less pronounced probably because our datasets are substantially larger than the publicly available datasets.

For the experiments with our production datasets we are using the \Bezier curve inputs which perform slightly better in terms of accuracy than the raw input encoding but are much faster to train and evaluate because of the shorter sequence lengths.

\section{System Performance and Discussion}
\label{sec:performance}

The setup described throughout this paper that obtained the best results relies on input processing with \Bezier spline interpolation (Sec.~\ref{sec:curves}), followed by 4--5 layers of varying width bidirectional LSTMs, followed by a final softmax layer. For each script, we experimentally determined the best configuration through multiple training runs.

We performed an ablation study with the best configurations for each of the six most important scripts\footnote{For Latin script we report results for 3 languages.} by number of users and compare the results with our previous work \cite{Google:HWRPAMI} (Table~\ref{tab:english_improvements}). The largest relative improvement comes from the overall network architecture stack, followed by the use of the character language model and the other feature functions.

\begin{figure*}
    \centering
    \includegraphics[width=0.7\linewidth]{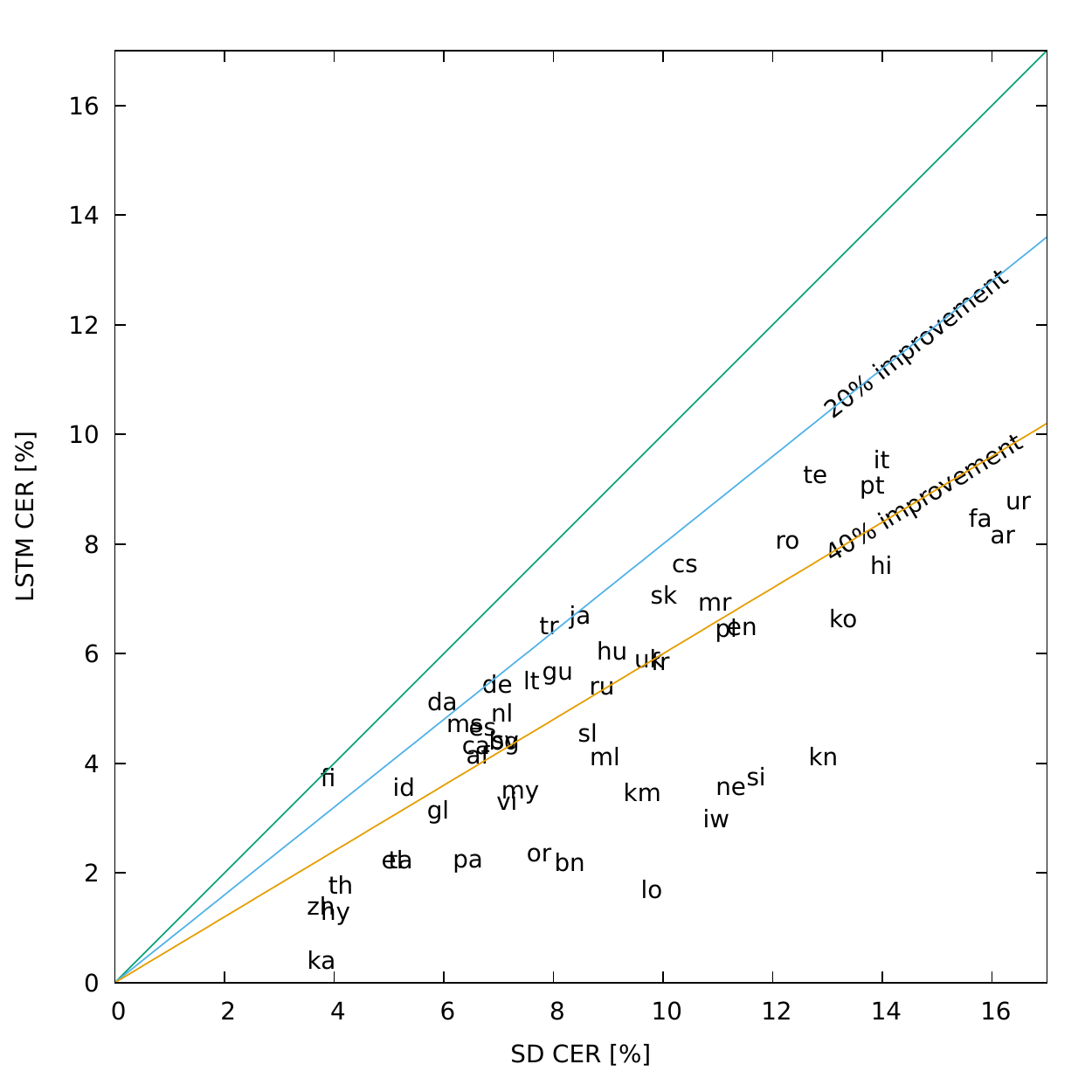}
    \caption{A comparison of the CERs for the LSTM and SD (Segment-and-Decode) system for all languages on our internal test sets with more than 2000 items. The scatter plot shows the ISO language code at a position corresponding to the CER for the SD system (x-axis) and LSTM system (y-axis). Points below the diagonal are improvements of LSTM over SD. The plot also shows the lines of 20\% and 40\% relative improvement.}
\label{fig:scatter}
\end{figure*}

In addition, we show the relative improvement in error rates on the languages for which we have evaluation datasets of more than 2\,000 items (Figure~\ref{fig:scatter}). The new architecture performs between 20\%--40\% (relative) better over almost all languages.

\subsection{Differences Between \iamondb, \IBMUB and our internal datasets}

\begin{table}
  \centering
  \caption{CER comparison when training and evaluating \iamondb, \IBMUB and our Latin training/eval set.
  We want to highlight the fundamental differences between the different datasets.}
  \label{tab:iamibmlatin}
  \begin{tabular}{cccc}
    train/test & \iamondb & \IBMUB & own dataset \\
    \hline
    \iamondb      &  3.8 & 17.7 & 31.2 \\
    \IBMUB        & 35.1 &  \textbf{4.1} & 32.9 \\
    own dataset   &  \textbf{3.3} &  4.8 &  \textbf{8.7} \\
    \hline
\end{tabular}
\end{table}

To understand how the different datasets relate to each other, we performed a set of experiments and evaluations with the goal of better characterizing the differences between the datasets.

We trained a recognizer on each of the three training sets separately, then evaluated each system on all three test sets (Table~\ref{tab:iamibmlatin}). The neural network architecture
is the same as the one we determined earlier (5 layers bidirectional LSTMs of 64 cells each) with the same feature
functions, with weights tuned on the corresponding tuning dataset. The inputs are processed using
\Bezier curves.

To better understand the source of discrepancy when training on \iamondb and evaluating on \IBMUB,
we note the different characteristics of the datasets:
\begin{itemize}
    \item{\IBMUB has predominantly cursive writing, while \iamondb has mostly printed writing}
    \item{\IBMUB contains single words, while \iamondb has lines of space-separated words}
\end{itemize}

This results in models trained on the \IBMUB dataset not being able to predict spaces as they are not
present in the dataset's alphabet. In addition, the printed writing style of \iamondb makes recognition harder
when evaluating cursive writing from \IBMUB. It is likely that the lack of structure through words-only data
makes recognizing \iamondb on a system trained on \IBMUB harder than vice-versa.

Systems trained on \IBMUB or \iamondb alone perform significantly worse on our internal datasets, as our
data distribution covers a wide range of use-cases not necessarily relevant to, or present, in the
two academic datasets: sloppy handwriting, overlapping characters for handling writing on small
input surfaces, non-uniform sampling rates, and partially rotated inputs.

The network trained on the internal dataset performs well on all three datasets. It performs
better on \iamondb than the system trained only thereon, but worse for \IBMUB. We believe
that using only cursive words when training allows the network to better learn the sample characteristics, than when learning about space separation and other structure properties not present in \IBMUB.

\begin{table*}
  \centering
  \caption{Character error rates on the validation data using
    successively more of the system components described above for
    English (en), Spanish (es), German (de), Arabic (ar),
    Korean (ko), Thai (th), Hindi (hi), and Chinese (zh) along with the respective
    number of items and characters in the test and training datasets. Average latencies for all languages and models were
    computed on an Intel Xeon E5-2690 CPU running at 2.6\,GHz.}
  \label{tab:english_improvements}
  \begin{tabular}{rlcccccccc}
    \multicolumn{2}{l}{Language}                     &    \sf en     & \sf es & \sf de & \sf ar & \sf ko & \sf th & \sf hi & \sf zh \\
    \hline
    \multicolumn{2}{l}{Internal test data (per language)}\\
    & {Items}        & ~$32\,645$  & ~$7\,136$ & $14\,408$ & $11\,617$ & $22\,951$ & ~$23\,608$ & ~$9\,030$ & $197\,547$ \\
    & {Characters}   & $162\,367$ & $40\,302$ & $83\,231$ & $84\,017$ &  $55\,654$ & $109\,793$  & $36\,726$ & $312\,478$ \\ 
    \\
    \multicolumn{2}{l}{Internal training data (per script)}\\
    & Items                       & \multicolumn{3}{c}{3\,293\,421}  &   570\,375    &  3\,495\,877 & 207\,833 & 1\,004\,814     &  5\,969\,179     \\
    & Characters                  & \multicolumn{3}{c}{15\,850\,724} & 4\,597\,255   &  4\,770\,486 & 989\,520 & 5\,575\,552  & 7\,548\,434      \\
    & Unique supported characters & \multicolumn{3}{c}{295}          & 337           &    3524      &  195     &   197        & 12\,726 \\
    
    \hline
    \\

    \multicolumn{2}{l}{System} & \multicolumn{7}{c}{CER [\%]}\\ \hline
      &  Segment-and-Decode \cite{Google:HWRPAMI}    &  7.5  &  7.2  &  6.0  & 14.8  & 13.8  &  4.1  & 15.7  & 3.76 \\
      & {BLSTM (comparison) \cite{Google:HWRPAMI}}    & 10.2  & 12.4  & 9.5   & 18.2  & 44.2  &  3.9  & 15.4  & ---  \\
    \hline

    & Model architecture (this work)& \multicolumn{3}{c}{5$\times$224} & 5$\times$160 & 5$\times$160 & 5$\times$128 & 5$\times$192 & 4$\times$192 \\
   
    (2) &  BLSTM-CTC Baseline Curves                  & 8.00  & 6.38  & 7.12  & 12.29 & 6.87  & 2.41  & 7.65  & 1.54    \\
    (3) &   + n-gram LM                              & 6.54  & 4.64  & 5.43  & 8.10  & 6.90  & 1.82  & 7.00  & 1.38    \\
    (4) &   + character classes                      & 6.60  & 4.59  & 5.36  & 7.93  & 6.79  & 1.78  & 7.32  & 1.39    \\
    (5) &   + word LM                                & 6.48  & 4.56  & 5.40  & 7.87  &  ---  &  ---  & 7.42  & ----    \\
    \hline
    \\
    
    \multicolumn{2}{l}{Avg. latency per item} & \multicolumn{7}{c}{[ms]} \\ \hline
      &  Segment-and-Decode \cite{Google:HWRPAMI}    &  315 & 359 & 372 & 221  & 389  &  165  & 139  & 208 \\
      &  This work    & ~23 & ~25 & ~26 & ~14  & ~20  &  ~13  & ~19  & ~30 \\ \hline
    \\
    
    \multicolumn{2}{l}{Number of parameters (per script)} & \multicolumn{7}{c}{} \\ \hline
      &  Segment-and-Decode \cite{Google:HWRPAMI}     &  \multicolumn{3}{c}{5\,281\,061}  & 5\,342\,561  & 8\,381\,686  &  6\,318\,361  & 9\,721\,361  & --- \\
    & This work & \multicolumn{3}{c}{5\,386\,170} & 2\,776\,937 & 3\,746\,999 & 1\,769\,668 & 3\,927\,736 & 7\,729\,994 \\ \hline

 \end{tabular}
\vspace{-1ex}
\end{table*}

\section{Conclusion}
\label{sec:conclusion}

We describe the online handwriting recognition system that we currently use at Google for  102 languages in 26 scripts. The system is based on an end-to-end trained neural network and replaces our old Segment-and-Decode system. Recognition accuracy of the new system improves by 20\% to 40\% relative depending on the language while using smaller and faster models. We encode the touch inputs using a \Bezier curve representation which performs at least as well as raw touch inputs but which also allows for a faster recognition  because the input sequence representation is shorter.

We further compare the performance of our system to the state of the art on publicly available datasets such as \iamondb, \IBMUB, and CASIA and improve over the previous best published result on \iamondb.

\subsubsection*{Acknowledgements}

We would like to thank the following contributors for fruitful discussions, ideas, and support: Ashok Popat, Yasuhisa Fujii, Dmitriy Genzel, Jake Walker, David Rybach, Daan van Esch, and Eugene Brevdo. 
We thank Google's OCR team for the numerous collaborations throughout the years that have made this work easier, as well as the speech recognition and machine translation teams at Google for tools and support for some of the components we use in this paper.

\bibliographystyle{spmpsci}

\end{document}